\documentclass[letterpaper, 10 pt, conference]{ieeeconf}  

\IEEEoverridecommandlockouts                              

\usepackage{graphics} 
\usepackage{graphicx} 
\usepackage{subfigure}
\usepackage{amsmath} 
\usepackage{amssymb}  
\usepackage{pifont}
\usepackage{algorithm, algorithmic} 
\usepackage{bm}
\usepackage{hyperref}
\usepackage{dblfloatfix}
\usepackage{multirow}
\usepackage{cite} 
\usepackage{array} 
\usepackage{wrapfig} 

\newcommand{\tflag}[1]{}
\newcommand{\tfflag}[1]{}

\usepackage[dvipsnames]{xcolor}
\title{\LARGE \bf
Non-Prehensile Aerial Manipulation using Model-Based Deep Reinforcement Learning
} 
\author{Cora A. Dimmig$^{1}$
and Marin Kobilarov$^{1}$
\thanks{$^{1}$Department of Mechanical Engineering and the Laboratory for Computational Sensing and Robotics (LCSR), Johns Hopkins University, Baltimore, MD 21218, USA. Email: {\tt\small cdimmig@jhu.edu, marin@jhu.edu}}%
}

\begin{document}

\maketitle
\thispagestyle{empty}
\pagestyle{empty}

\begin{abstract}

With the continual adoption of Uncrewed Aerial Vehicles (UAVs) across a wide-variety of application spaces, robust aerial manipulation remains a key research challenge. Aerial manipulation tasks require interacting with objects in the environment, often without knowing their dynamical properties like mass and friction \textit{a priori}. Additionally, interacting with these objects can have a significant impact on the control and stability of the vehicle. We investigated an approach for robust control and non-prehensile aerial manipulation in unknown environments. In particular, we use model-based Deep Reinforcement Learning (DRL) to learn a world model of the environment while simultaneously learning a policy for interaction with the environment. We evaluated our approach on a series of push tasks by moving an object between goal locations and demonstrated repeatable behaviors across a range of friction values.

\end{abstract}
\section{Introduction}

Robust aerial manipulation can enable a broad spectrum of operational scenarios, including enabling tasks that are impossible, dangerous, or costly in time and/or resources for humans to complete.
Consequently, the adoption of Uncrewed Aerial Vehicles (UAVs) with manipulation capabilities, as discussed in \cite{OlleroFranchi2022Past}, has increased in many application spaces such as parcel delivery, warehouse management, and sample collection. 

A key challenge in aerial manipulation is accomplishing a task in the presence of occlusions or with dynamic environment interactions. For example, in agricultural applications, in order to reach a piece of fruit, the vehicle may need to push past surrounding foliage. In cluttered environments, objects may need to be pushed out of the path to another object in order to view a target of interest. 
These types of non-prehensile environment interactions can be challenging, if not impossible to define \textit{a priori}.
Classical control approaches typically require a precise geometric model and dynamic state of each movable object in the environment, which frequently does not scale to the case of foliage or clutter. Modeling of interactions in such complex environments is extremely challenging due to unknown inertial properties and complex contact dynamics. 
The work herein considers non-prehensile environment interactions with a dynamic, underactuated vehicle in a way that is easily extensible to a wide class of objects and enables dynamic contact with the environment. 

\begin{figure}[tbh]
    \vspace*{2mm}
    \centering
    \includegraphics[width=1.0\columnwidth]{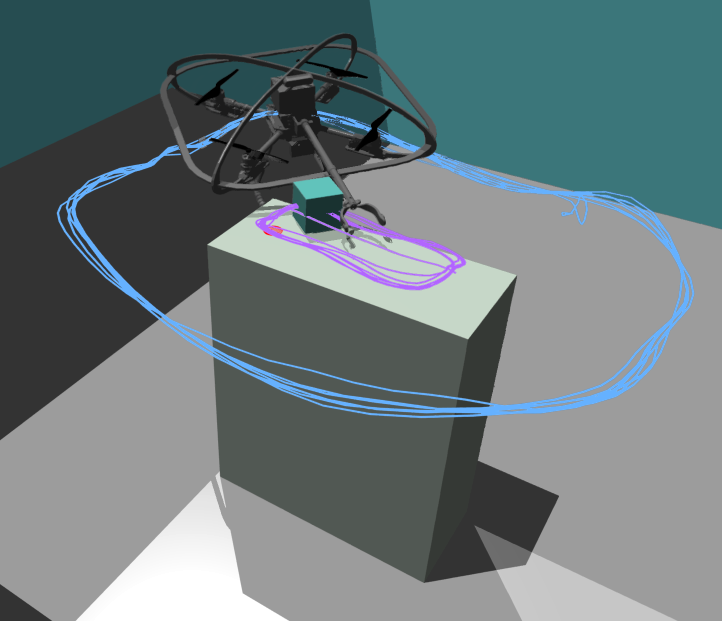} 
    \caption{Aerial vehicle pushing an object between goal points on alternating ends of the table in simulated environment. Object's trajectory is shown in purple and vehicle's trajectory is shown in blue.}
    \label{fig:back-and-forth}
\end{figure}

\begin{figure}[tbh]
    \centering
    \subfigure[Simulated Vehicle]
        {\includegraphics[width=0.44\columnwidth]{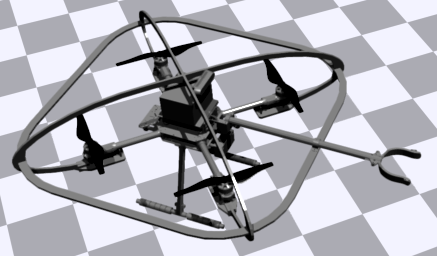} \label{fig:quad_simulation}} 
    \subfigure[Hardware Vehicle]
        {\includegraphics[width=0.51\columnwidth]{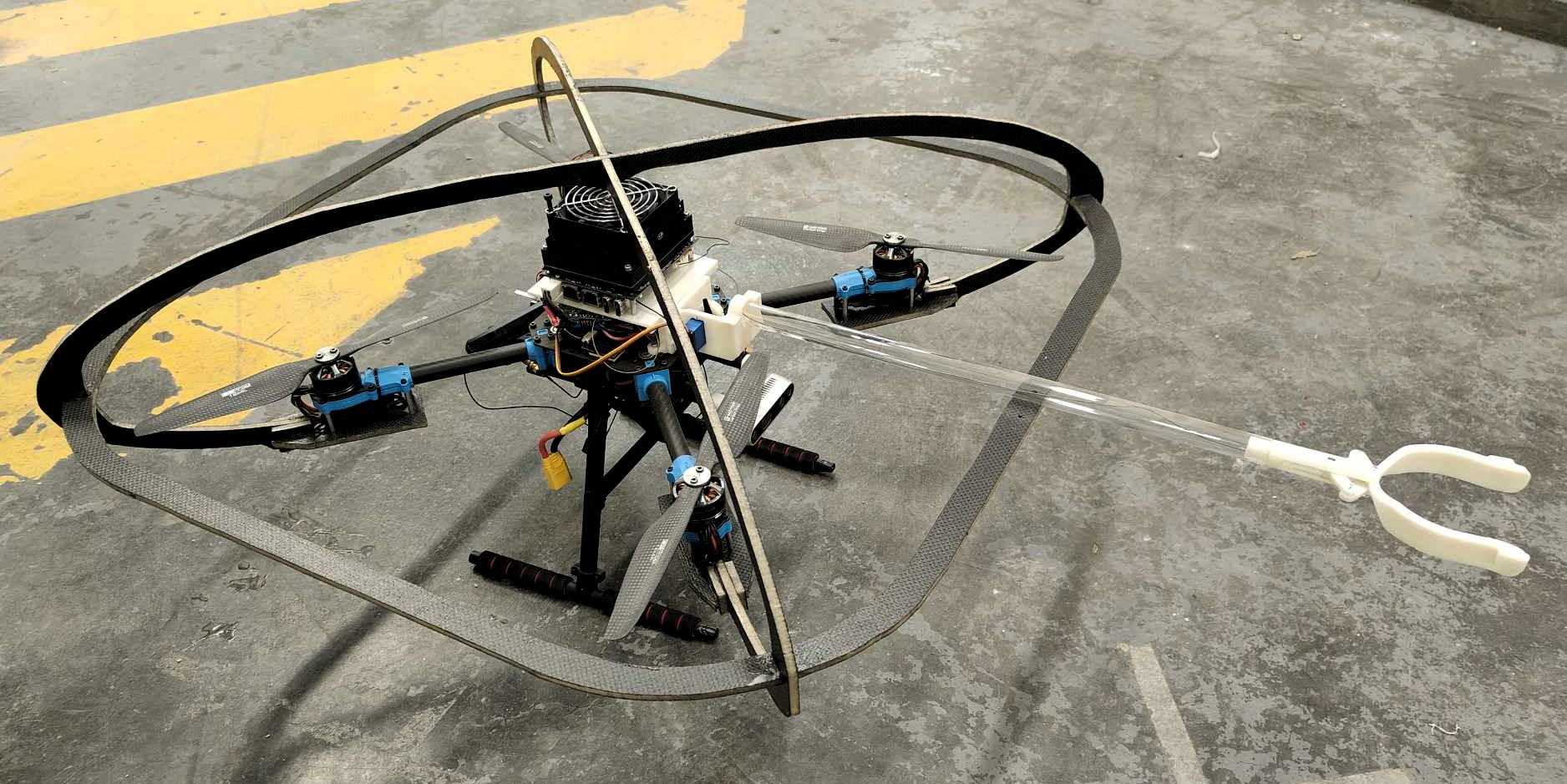}
        \label{fig:quad_hardware}} 
    \caption{Aerial manipulation platform with a fixed gripper for non-prehensile and grasping tasks.}
    \label{fig:platform}
\end{figure}

\begin{figure}[tbh]
    \vspace*{2mm}
    \centering
    \includegraphics[width=\columnwidth]{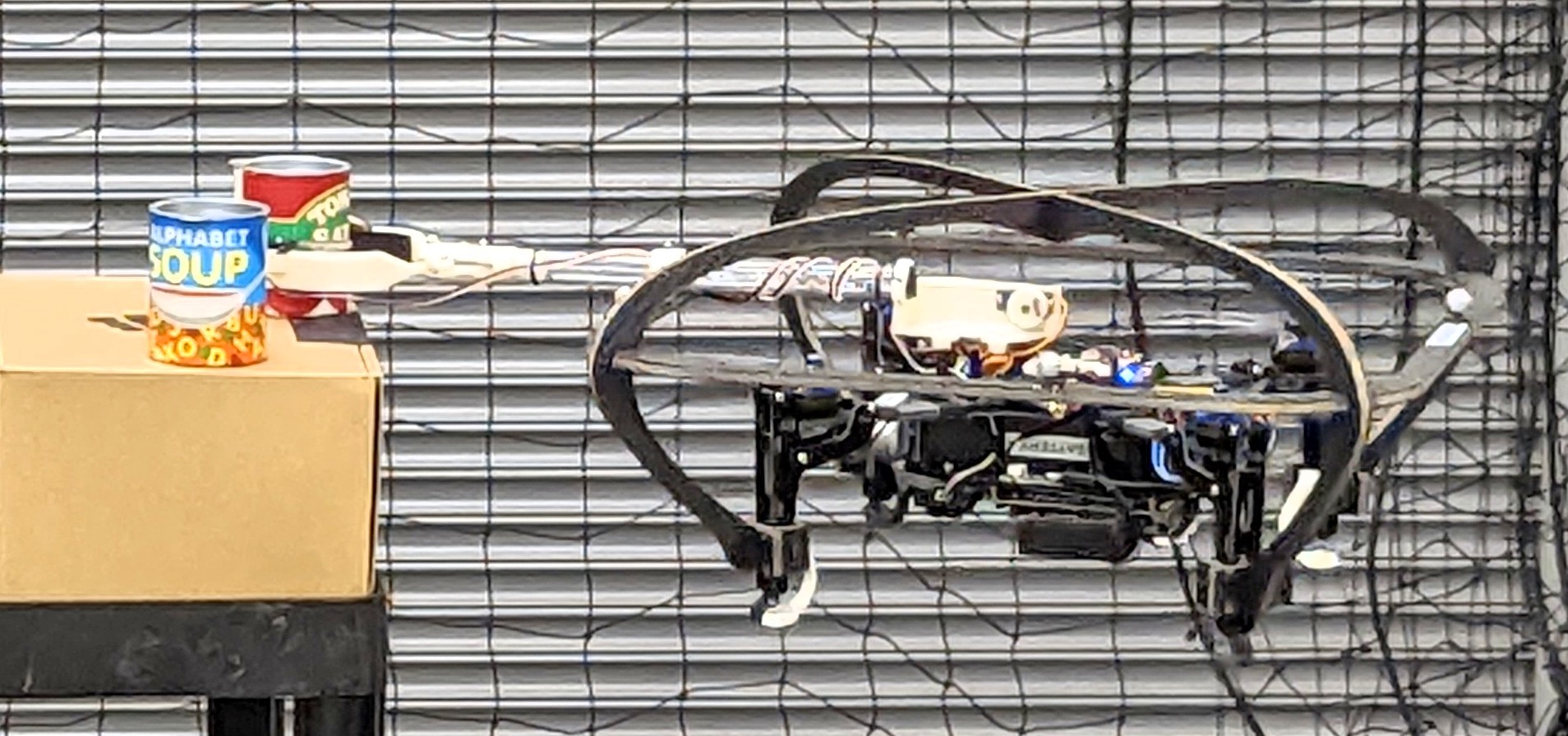}
    \caption{
    Example aerial manipulation experimental scenario. The vehicle started approaching perpendicular to the scene and a traditional controller required the vehicle to move to the side of the scene to grasp the tomato sauce can. In this work, we explore capabilities to more robustly accomplish this task, for example, in cases where the side may be inaccessible the alphabet soup would need to be pushed in order to reach the target object.}
    \label{fig:obstruction}
\end{figure}

More specifically, we consider the question: how can an aerial vehicle reliably position a given object in a desired goal location, without explicitly knowing and modeling its dynamical properties like mass or friction? Instead, the robot is expected to implicitly understand the contact dynamics with the environment by visually observing its interaction.

We propose a Deep Reinforcement Learning (DRL) approach that tightly couples planning, visual perception, and dynamic reasoning to enable planning for contact with the environment.
In particular, we present results adapting the Model-Based DreamerV3 algorithm \cite{Hafner2023} to the challenging case of floating-base manipulation.
We consider test scenarios involving pushing objects across a surface to analyze the capability to learn behaviors that account for varying dynamical properties. We considered variable friction across the surface in addition to variable goal locations, thus requiring the agent to jointly 
optimize over the navigation and manipulation tasks. Fig.~\ref{fig:back-and-forth} depicts the trajectory of the object of interest in a scenario where the vehicle must push the object back and forth between goal points at alternating ends of the table. 

We formulate our state, action, and reward spaces with considerations for transfer to hardware vehicles in the future. Fig.~\ref{fig:platform} shows the analogous hardware platform to our simulated vehicle. 
Fig.~\ref{fig:obstruction} depicts an example experimental scenario that could be performed efficiently by the capability proposed herein. An obstruction in the field of view of the vehicle generally requires the vehicle to maneuver around the obstruction entirely, and the obstruction may disrupt detecting the target object.  The ability to reason about these scenarios and push obstructions out of the way could significantly improve the robustness of autonomous system planning. 
Additionally, we believe our approach will be easily extensible to a wide variety of platforms and scenarios for both non-prehensile and prehensile aerial manipulation.

\section{Contributions}

The objective of this paper is to introduce a new approach to planning for non-prehensile manipulation in unknown environments. Our contributions include:
\begin{itemize}
    \item Introduction of a generalized DRL pipeline for training numerous agents simultaneously in simulation with varying contact dynamics
    \item A novel method to learn a policy for robust control and non-prehensile aerial manipulation, including the development of specialized state space, action, and reward structures 
    \item Experimental results demonstrating the robustness of our proposed algorithm across a wide range of friction values, including analysis and insights for further study
\end{itemize}

\section{Related Work}

\subsection{Navigation and Planning}

Robot navigation and mobile manipulation is commonly addressed through hand-engineered or sampled waypoints.
In \cite{Dimmig2023a}, a small form factor vehicle was used for pick-and-place experiments with entirely onboard computation, however the control strategy used a series of waypoints that were defined relative to the detected object of interest, which can be limiting in the presence of occlusions and obstructions. 
Additionally, finding dynamically reachable points is increasingly difficult in clutter, and common methods often inherently disallow contact with the environment \cite{Kulkarni2024} though contact may be essential for reaching an optimal observation position or object of interest. Most commonly, approaches rely on building maps of the environment \cite{OlleroFranchi2022Past}, which can be confounded by localization, sensing, and matching errors, in addition to adding latency to the system. 
Penicka et. al \cite{Penicka2022} demonstrated an approach for planning fast minimal-time trajectories through a drone racing course using DRL, however the approach relied on a known map with specified waypoints which is generally impractical in real-world scenarios. 
DRL approaches enable planning for contact with the environment, however many end-to-end DRL approaches \cite{Sun2022} are not sample efficient and thus require prohibitive training times and quantities of data.
We propose a DRL method, designed to be sample efficient, that allows contact with the environment and is meant to operate without any prior information about the environment.

There is a significant divide in approaches to navigation that comes from differences between optimal control and learning-based approaches. 
Song et. al. \cite{Song2023} investigated the performance differences between optimal control and RL on autonomous drone racing tasks. They concluded that the RL method outperformed optimal control in part due it's ability to optimize a task-level objective rather than decomposing the problem into planning and control steps. This can be essential when considering simultaneously navigating and performing a task, as in the aerial manipulation domain.  

A key challenge with learning-based approaches comes from transferring policies trained in simulation to the real world. However, significant progress has been made to this end. 
In \cite{Loquercio2021}, Loquercio et. al. demonstrated zero-shot transfer of policies trained in simulation, using imitation learning with a privileged expert, to complex real-world environments by simulating realistic sensor noise.
Similarly using realistic sensor behaviors in simulation, Kulkarni et. al. \cite{Kulkarni2024} demonstrated collision-free flight in both simulation and hardware using a deep collision encoding trained on both simulated and real depth images. 

\subsection{Autonomous Push Task}

In this work, we investigate non-prehensile manipulation of an object of interest.
Push tasks can enable further manipulation of objects, such as by clearing clutter from an environment to be able to grasp an object of interest, as explored in \cite{Zeng2018}. 
The study of non-prehensile manipulation has long relied on modeling of friction in the environment. However, these friction models frequently assume uniform friction distributions \cite{Mason1986} (which does not hold in practice) or can be highly specialized \cite{Bauza2017, Liu2023} (which can be impractical to measure and compute in many applications). 
This has motivated the use of DRL for pushing applications, often studied with fixed robotic arms, as in \cite{Zeng2018, Peng2018}.
Researchers have explored dynamics randomization with LSTMs in simulation to aid in transferring policies to real systems for both learning to push objects \cite{Peng2018} and predicting the sliding motion of objects with unknown parameters \cite{Cong2020}.
We designed our method around being robust to varying friction in the environment when the friction cannot be predicted prior to interaction. 

The authors of \cite{Driess2022} investigated RL with Neural Radiance Fields (NeRFs) to utilize the strong 3D inductive bias properties of NeRFs. They considered a planar pushing task, though their solution required multiple camera views, object masks, and pretraining of the latent representation offline. 
In this work, our encoder and decoder are trained online with only one camera view of depth information.

\subsection{Aerial Vehicle Environment Interaction}

Many forms of aerial manipulation have been studied \cite{OlleroFranchi2022Past}, including for momentary, loose, and strong coupling with the environment \cite{Orsag2017}.
Pick-and-place tasks are by far the most common. The work in \cite{Dimmig2023a} investigated grasping objects of interest using onboard detection and then placing them at a detected destination location. 

Aerial manipulation push tasks are less common in the literature and have been most frequently investigated for pushing open doors or large objects, which requires considering the forces between the structure and the vehicle. For this purpose, researchers have used the lift of the propellers to generate sufficient force \cite{Tsukagoshi2015} or have considered the relationship between interaction forces and stability of the vehicle \cite{Lee2021}.
Additionally, researchers have investigated the forces inflicted on the vehicle due to interaction with the environment with Force-Torque sensors \cite{Malczyk2023}. We investigated non-prehensile aerial manipulation in this work without any specialized instrumentation, considering continued contact between a UAV and the environment to achieve an objective.


\section{Aerial Interaction Technical Approach}

Our primary objective of this work was to investigate UAV interaction with the environment. Toward this goal, we introduce a generalized DRL pipeline to efficiently train an agent in simulation for complex navigation and interaction tasks. We evaluate this pipeline in a scenario where the aerial vehicle must push a target object to as many goal positions on a table as possible within a set time frame. As a form of domain randomization, we consider a range of friction values between the object and table, requiring the simulated agent to learn to adapt to each environment. Additionally, we evaluate the performance of our algorithm in varied scenarios. 
We hypothesized that the agent would learn to perform cautious actions in complex environments to accomplish the intended goal while adapting to the particular environment, without any form of intermediate map representation of the environment.

\begin{figure}[tbh]
    \vspace*{2mm}
    \centering
    \includegraphics[width=0.95\columnwidth]{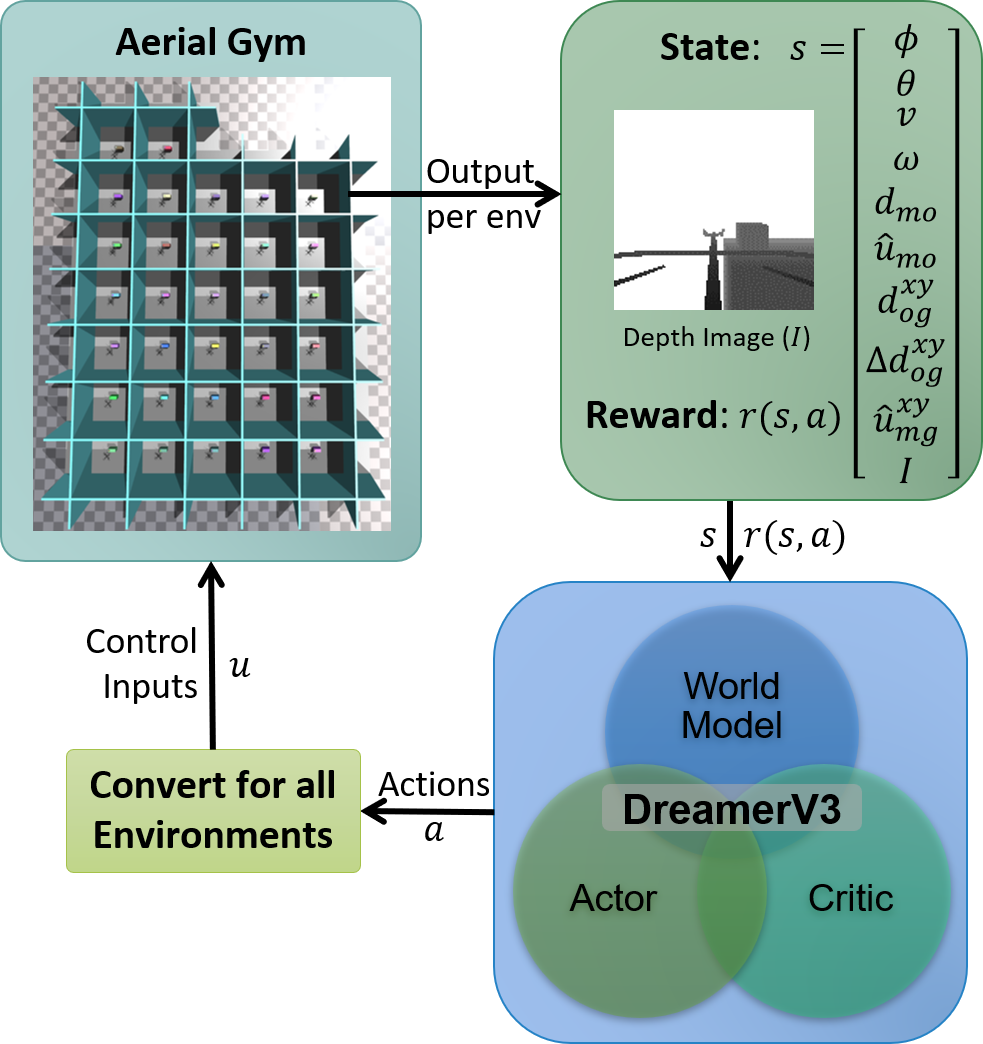} 
    \caption{Algorithmic architecture for model-based deep reinforcement learning in simulated Aerial Gym environments.}
    \label{fig:architecture}
\end{figure}

\subsection{Background}

\subsubsection{DreamerV3}

The foundation of our algorithmic approach is the DreamerV3 algorithm developed by Hafner et. al. \cite{Hafner2023}. 
The authors in \cite{Hafner2023} demonstrated DreamerV3's ability to adapt to diverse domains with a single set of hyperparameters, showing performance as a general algorithm for robot locomotion, navigation, Atari, and 3D domains such as maze environments and Minecraft. 
The DreamerV3 algorithm learns a world model from perception data. Sensory inputs are encoded and the world model learns to project the latent state of these inputs forward to predict future outcomes of potential actions. This allows the model to train by predicting forward the state of the world (i.e., by imagining) in addition to new inputs, making the algorithm highly sample efficient. An Actor-Critic learning approach is trained simultaneously with the world model. The critic judges the value of each scenario and the actor learns to reach the valuable scenarios. 
An earlier iteration of the DreamerV3 algorithm was demonstrated for physical robot learning, with small amounts of real-world interaction, in their algorithm DayDreamer \cite{Wu2023}. The world model's ability to predict outcomes of potential actions is shown to significantly reduce the overall training time on the real robot. In \cite{Wu2023}, the author's demonstrated performance for a quadruped learning to walk, pick and place experiments with fixed arms, and a 2D visual navigation task.

To the best of our knowledge, the work proposed herein is the first use case of the DreamerV3 algorithm on a flying vehicle and, in particular, for aerial manipulation tasks. 
We hypothesized that Dreamer's ability to predict into the future will be highly valuable for the combined navigation and manipulation task. Additionally, the sample efficiency allows for future extensions of this work with onboard training. 
In this work, we adapted a PyTorch implementation of DreamerV3 \cite{dreamerv3torch}.

\subsubsection{Aerial Gym}

\begin{figure}[tbh]
    \centering
    \includegraphics[width=0.75\columnwidth]{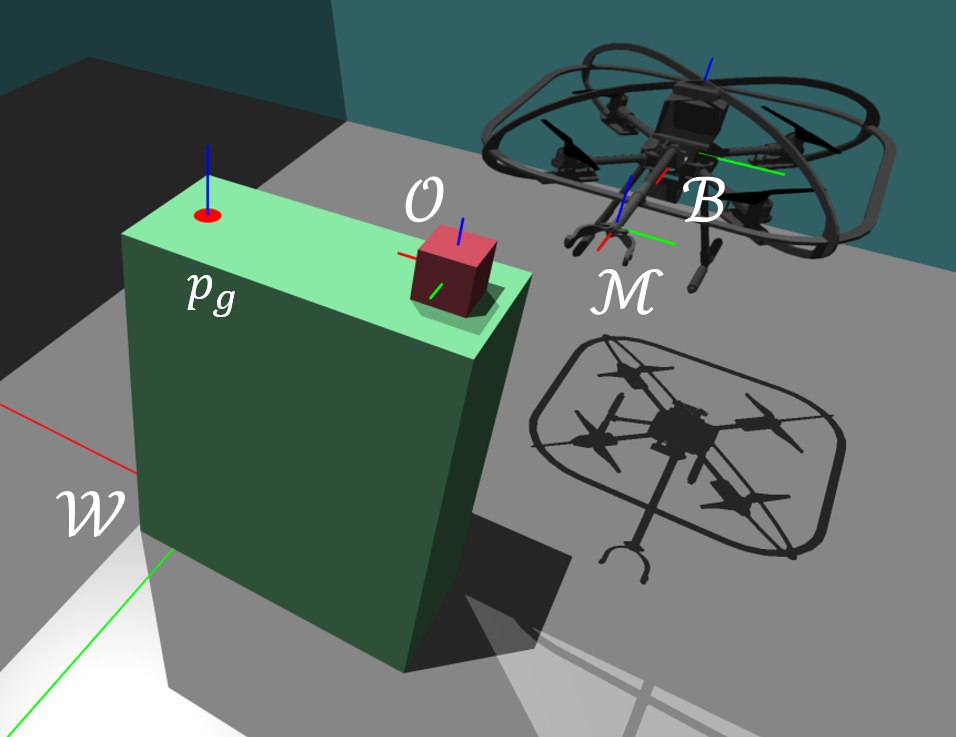} 
    \caption{Simulation environment for baseline experiments.}
    \label{fig:env}
\end{figure}

We are using NVIDIA's Isaac Gym \cite{Makoviychuk2021} with the Aerial Gym extension \cite{Kulkarni2023} for simulation of an aerial manipulation platform. 
We selected Isaac Gym due to it's ability to simulate a large number of environments in parallel on the GPU \cite{Dimmig2023Survey}.
We developed a middleware to interface Aerial Gym's parallelized environment with DreamerV3's environment instance based architecture. A general depiction of our algorithmic architecture is shown in Fig.~\ref{fig:architecture}. In Aerial Gym we create environments as seen in Fig.~\ref{fig:env} with the quadcopter model from Fig.~\ref{fig:platform} in a collision tolerant carbon fiber foam cage and gripper extension package from \cite{Dimmig2023a}, an object of interest, and a structure that the object sits on. The simulated vehicle is a replica of the existing hardware vehicle in Fig.~\ref{fig:platform} and is not optimized for non-prehensile manipulation tasks. Optimizing the design, such as by varying the length of the arm or the contact surfaces, could overall increase the vehicle's potential for non-prehensile capabilities. 

Fig.~\ref{fig:env} depicts an example simulation environment for the baseline experiments and defines the key frames we consider and corresponding origin points $p$ which we define in the world frame. These include: the world frame $\mathcal{W}$ with origin $p_w$, the vehicle's base frame $\mathcal{B}$ with origin $p_b$, a frame $\mathcal{M}$ at the center of the exposed portion of the fixed manipulator with origin $p_m$, and the object's frame $\mathcal{O}$ with origin $p_o$. Additionally, the goal region (for the object's center to reach) is depicted in red and the center of this location is $p_g$. In Fig.~\ref{fig:env}, $p_g$ is along the z-axis in the world frame, this is not always the case and depends on the scenario setup, which will be further explained in Section~\ref{sec:exp}.
In this evaluation, we used simple $0.1~m$ cubes with variable friction coefficients as our objects of interest.

\subsection{RL Framework}

We based our RL framework on the DreamerV3 algorithm and used the standard RL paradigm: the vehicle, or agent, interacts with the environment based on a policy with the ultimate goal of maximizing a reward. 

At each environment time step, we define a state of the vehicle and environment $s \in \mathbb{S}$ (which we assume is fully observable), action $a \in \mathbb{A}$, and reward function $r(s,a): \mathbb{S} \times \mathbb{A} \rightarrow \mathbb{R}$. The agent aims to maximize the episodic return of this reward signal, i.e. $r(s,a)$ summed over all steps in an episode.  

\subsection{State Space}

We comprise our state $s$ with the following observations of the vehicle and environment state: 
\begin{itemize}
\item roll $\phi$ and pitch $\theta$ in $\mathcal{W}$, 
\item velocity $v$ and angular velocity $\omega$ in $\mathcal{B}$, 
\item a distance $d_{mo}$ and unit vector $\hat{u}_{mo}$ from the arm $p_m$ to the object of interest $p_o$ in $\mathcal{M}$, 
\item a planar distance $d_{og}^{xy}$ from the object $p_o$ to the goal point $p_g$, 
\item the change in planar distance $\Delta d_{og}^{xy}$ from the object to the goal between the last step and the current step,
\item a unit vector $\hat{u}_{mg}^{xy}$ from the arm $p_m$ to the goal point $p_g$ along the $x$ and $y$ axes of $\mathcal{M}$,
\item and depth images $I$. 
\end{itemize}

Ultimately we intend our results to be transferable to hardware vehicles in the future, so we select depth images rather than RGB images to be more robust to the differences between simulation and reality, as demonstrated in \cite{Loquercio2021}.
Additionally, we define unit vectors as a normalized signal that indicates the direction the vehicle should first move in to reach the object with the arm, $\hat{u}_{mo}$, and then to push the object to the goal, $\hat{u}_{mg}^{xy}$. The distances $d_{mo}$ and $d_{og}$ represent the magnitude of these maneuvers. Notably, we define the unit vector for pushing the object relative to $\mathcal{M}$ since this frame is fixed relative to the vehicle versus $\mathcal{O}$ will rotate with the object. However we utilize the distance $d_{og}$ relative to $\mathcal{O}$ since this is the distance we want to minimize directly to accomplish the push task.

\subsection{Action Space}

We command the simulated vehicle using a parallelized velocity controller on SE(3) implemented in Aerial Gym \cite{Kulkarni2023}. 
Our control inputs are a desired velocity vector $v_d \in \mathbb{R}^3$ and yaw-rate $\omega_{z_d}$ in the vehicle's frame. 
We calculate these control inputs 
using the actions output from our agent $a = [-1,1]^4$. We designed the elements of $a$ to correspond to the speed in the $xy$-plane relative to the vehicle's frame $\mathcal{B}$, yaw about the $z$ axis in the global frame $\mathcal{W}$, speed in the vehicle's $z$ direction, and yaw-rate relative to the vehicle. We transform the outputs $a$ to our control input $u= [v_d; \omega_{z_d}]$ at each time step. This formulation was inspired by \cite{Kulkarni2024}.
\begin{align}
    v_{x_d} &= s_{xy,m} \bigg( \frac{a_1 + 1}{2} \cos(\theta_m a_2) \bigg) \\
    v_{y_d} &= s_{xy,m} \bigg( \frac{a_1 + 1}{2} \sin(\theta_m a_2) \bigg) \\
    v_{z_d} &= s_{z,m} a_3 \\
    \omega_{z_d} &= \omega_{z,m} a_4
\end{align}

We bound the maximum speeds $s_{xy,m}$ and $s_{z,m}$, yaw $\theta_{m}$, and yaw-rate $\omega_{z,m}$. In this evaluation, we use the values $1~m/s$, $0.5~m/s$, $\pi~rad$, and $\frac{\pi}{4}~rad/s$, respectively. This formulation assures the vehicle can only traverse in directions that are viewable from the onboard camera, e.g. the vehicle cannot move backwards into space it cannot see. 

\subsection{Navigation and Task Reward Structure}

\subsubsection{Reward Functions}
\label{sec:rew_func}

We define reward functions for each step of an episode and then minimize over the total magnitude of the rewards across the entire episode. 
We use the following general function for our positive reward terms.
\begin{align}
    f^+(d, \gamma) = \frac{\gamma}{1 + d^2}
\end{align}
Here $d$ is a distance 
and $\gamma > 0$ is the magnitude of the reward. The value of $f^+(d, \gamma)$ increases as the distance $d$ decreases, in particular, $f^+(d, \gamma)$ increases more rapidly as $d$ approaches zero (i.e., an increased incentive as $d$ decreases). 

We then define the following function for negative reward terms $f^-(d, \gamma, \tau)$. Similarly this function $f^-(d, \gamma, \tau)$ increases as the distance $d$ decreases, however the magnitude is always negative up to a maximum value of $-1$. Additionally, we include a threshold $\tau$ such that distances below this threshold are given the maximum value.
\begin{align}
    f^-(d, \gamma, \tau) = \begin{cases}
        f^+(d, \gamma) - \gamma - 1, & d > \tau \\
        -1, & d \leq \tau
    \end{cases}
\end{align}

Next, we define an incremental reward function based on a change in distance. This reward will be positive or negative based on the sign of $\Delta d$, where $\Delta d$ represents the change in distance between the last time step and the current time step. Thus, this reward is positive when the distance is decreasing.
\begin{align}
    f^{\Delta}(\Delta d, \gamma) = \gamma \Delta d
\end{align}

Lastly, we include a sparse impulse reward function, such as for completing the overall task, $f^{|}(d, \gamma, \tau)$. When within a specified distance threshold $\tau$, a reward $\gamma$ is received. 
\begin{align}
    f^{|}(d, \gamma, \tau) = \begin{cases}
        \gamma, & d < \tau \\
        0, & d \geq \tau
    \end{cases}
\end{align}

\subsubsection{UAV Navigation and Object Manipulation Rewards}

We formulate our specific reward terms for our navigation and manipulation task to incentivize the agent to learn to move the fixed gripper arm toward the object of interest and to move the object toward its goal location. 

To define key distances, we use the reference frames defined in Fig.~\ref{fig:env}.
The origins of these frames are in $\mathbb{R}^3$. When we are referring to only particular components of a vector, we use superscripts representing the global $x,y,z$ directions. Using the standard L2-norm, we define the distances between the gripper center and object both in the plane ($xy$) and in the $z$ direction as $d_{mo}^{xy}$ and $d_{mo}^{z}$, respectively, and the distance between the object and goal in the plane as $d_{og}^{xy}$. Lastly, we define $d_{q}$ as a representation of the amount the vehicle is tilting. $d_{q}$ represents the distance in $z$ between the vehicle's body frame and the world frame using the unit vector $e_3 = [0,0,1]^T$ and the rotation matrix $R$ based on the vehicles orientation. 
\begin{align}
    d_{mo}^{xy} &= \lVert p_{m}^{xy} - p_o^{xy}\rVert \label{eq:dgo_xy}\\
    d_{mo}^{z} &= \lVert p_{m}^{z} - p_o^{z}\rVert \label{eq:dgo_z}\\
    d_{og}^{xy} &= \lVert p_{o}^{xy} - p_g^{xy}\rVert \label{eq:dot_xy}\\
    d_{q} &= \lvert 1 - e_3^T R(\phi, \theta) e_3 \rvert \label{eq:dtilt}
\end{align}

Using the reward functions we defined in Section~\ref{sec:rew_func}, we can express our overall reward function $r(s,a)$, which is a function of the components of the state $s$ and action $a$. In this case, we define our reward function without any components of the action space (i.e., $r(s,a) = r(s)$). Additionally, we include the values for the reward magnitudes $\gamma$ and distance thresholds $\tau$ in our formulation (based on our simulated vehicle and environment) that we used in our evaluation. 
In particular, we scale the navigation rewards by $2$, task rewards by $1000$, and impulse complete reward by $1500$. These values are selected based on each reward function's relative importance to our overall objective. 
\begin{align}
\begin{split}
    r(s,a) &= f^-(d_{mo}^{xy}, 2, 0.125) \Big( 1 + f^+(d_{q}, 1) \Big) \\
    &\quad\quad + f^-(d_{mo}^{z}, 2, 0.05)
    + f^{\Delta}(\Delta d_{og}^{xy}, 1000) \\
    & \quad\quad + f^{|}(d_{og}^{xy}, 1500, 0.025)
\end{split}
\label{eq:reward}
\end{align}

In (\ref{eq:reward}), we included the navigation rewards as negative terms using $f^-$ to distinguish them from the ultimate objective of pushing the object to an intended goal location. As negative terms, maximizing these rewards will never be as valuable as maximizing the ultimate objective terms (which are positive). The negative navigation terms include a reward for the distance from the gripper center to the object in $xy$ and $z$ (i.e., rewards $f^-(d_{mo}^{xy}, 2, 0.125)$ and $f^-(d_{mo}^{z}, 2, 0.05)$, respectively). We consider the planar motion and $z$ motion separately to be able to set different regions of success. In the plane, anywhere along the gripper can be aligned with the object to be able to push it, so we consider a larger region of $0.125~m$, since that is half the length of the exposed arm. In the $z$ direction we consider half the height of the object, $0.05~m$. We scale the reward for minimizing the vehicle's tilt ($f_r(d_{q}, 1)$) by the reward for the gripper's position such that the tilt reward becomes more important closer to the object. 

Finally, we include the task specific rewards as positive terms. First, $f^{\Delta}(d_{og}^{xy}, 1000)$ is a reward for the distance between the object and the goal decreasing. Second, $f_{|}(d_{og}^{xy}, 1500, 0.025)$ is a large impulse completion reward for the object being within a specified threshold of the target region. In this case, our region to consider a goal completed is when the object's center is within $0.025~m$ of the goal point (which is 1/4 the length of an edge of the object). The magnitude of this reward will always be greater than any other combination of rewards.  

\subsection{Domain Randomization}
\label{sec:domain_scenarios}

We assign different friction coefficients between the object and table in each training environment as a form of domain randomization. Our intention is for the agent to learn one policy that will generalize across the different environments, since friction values are infrequently known \textit{a priori}. 
As described in \cite{Peng2018}, since our world model is a recurrent model, system identification is implicitly embedded into the policy since the internal memory is a summary of past states and actions, which allows the policy to infer the dynamics of the system. 

\section{Experimental Results and Discussion}
\label{sec:exp}

We build experimental environments in constrained $5~m$ rooms with a simulated vehicle, table, and object of interest, as seen in Fig.~\ref{fig:env}. In each experiment we use a $0.1~m$ cube of mass $0.5~kg$ as the object of interest. 
We train the agent in 32 environments simultaneously, each with a different coefficient of friction between the object and table. The coefficients of friction are evenly distributed between the ranges $[0.05, 0.3]$ and $[0.55, 0.8]$.
We then evaluate performance with friction values between $0.2$ and $0.6$ since these are most common for objects on wood \cite{Ashby2010}. We train with a larger spectrum of friction coefficients for the system to be more robust to variations. We observed that focusing on the ends of the spectrum enabled the agent to develop a policy that was more robust to all possible friction values (since the higher and lower frictions are the most difficult to adapt to). 
Additionally, we found that when including the $0.05$ friction value (e.g., ice on wood) we saw an improvement in performance. We hypothesize that the significant increase in sliding of the object led to more reward variation at the beginning of training from which the agent could learn. 

We consider two tasks in our evaluation. In the first, the vehicle starts in a fixed position relative to the object and table, as depicted in Fig.~\ref{fig:env}. To maximize rewards, the vehicle needs to push the object to a goal point located on the opposite end of the table ($0.5~m$ away). Upon reaching that point, the goal updates to the other end of the table, and so forth for the duration of the episode. 
In our second task, we randomly generate goal locations across the surface of the table (minimally $0.15~m$ apart).

We train each environment for 100 seconds (1000 steps with a 0.1 time step). The environments automatically reset under any of the following conditions: at the end of the time limit, when the vehicle is in continuous collision with the environment for 2 seconds (including with the table, but not including the target object), when the object is no longer in contact with the table for 2 seconds, and if the vehicle leaves a 5 meter radius. 

When we evaluate performance, we do not reset the scenario when the vehicle collides with the table. We include these resets in the training to incentivize the vehicle to learn a policy without relying on touching the table since that could cause the vehicle to become unstable. However, in the real world the vehicle would be able to touch the table. 

We select the training step of the policies to use for evaluation based on performance (completing at least one goal across all environments). Additionally, we trained with two different random seed values in order to generalize the performance of our algorithm. We then tested the policies using a distinct seed value (not used during training). We evaluated 100 environments for each friction value of interest. We consider a goal to be complete when the distance from the object to the goal is less than $0.025~m$.

\begin{figure}[tbh]
    \centering
    \subfigure[Alternating goal points]
        {\includegraphics[width=0.95\columnwidth]{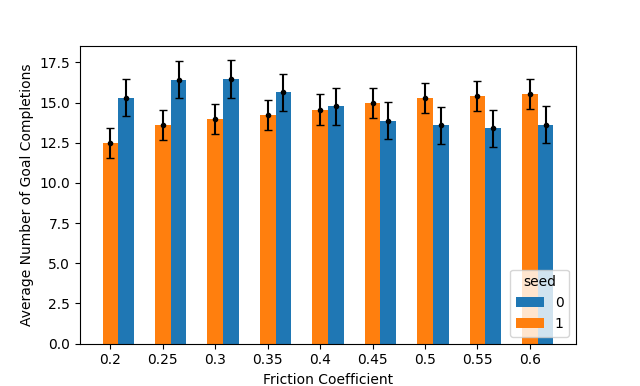} \label{fig:alternating}} 
    \subfigure[Random goal points]
        {\includegraphics[width=0.95\columnwidth]{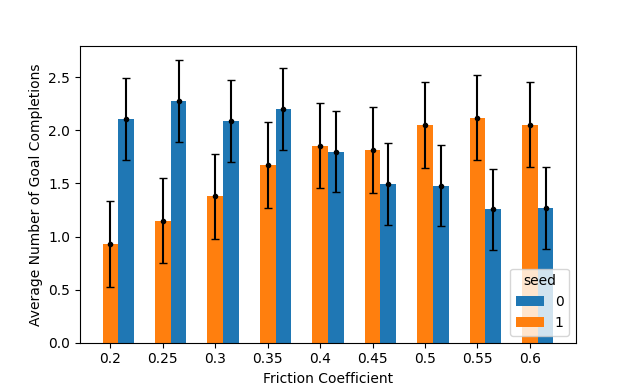}
        \label{fig:random}} 
    \caption{Average number of goals completed within 100 second episodes for two training seed values. Averaged across 100 environments for each friction value. Error bars are $\pm$ one standard deviation. }
    \label{fig:completions}
\end{figure}

Fig.~\ref{fig:completions} shows the average number of goal completions for the two task types. The alternating goal points is a highly repetitive task, which allows for the vehicle to find shortcuts. Ultimately, the vehicle learns to fly quickly around the table continually pushing to the right (regardless of if the object is being moved). This efficiently completes the task but does not demonstrate directly utilizing the input information (in particular, the position of the object relative to the goal). Additionally, the continuous rightward motion aids in the image prediction step since the goal locations on either side of the table are symmetrical when the vehicle switches sides as well, so the set of possible scenarios seen by the camera is reduced. In comparison, the average number of completions for the randomized goal locations is about a factor of 10 less, however still on average completing the task at least once in each environment across all friction values. The random goal locations is a significantly more challenging task since a repetitive circular motion (as demonstrated for the alternating goal points) is no longer sufficient and a more intentional method of utilizing the state information is necessary. 
We include a video demonstrating the vehicle performing both non-prehensile manipulation tasks\footnote{Video of Evaluation: \href{https://youtu.be/h5G1srG6H-o}{https://youtu.be/h5G1srG6H-o}}. Additionally, the video depicts the training progression of the agent learning to adapt to the differences in dynamic properties.

\addtolength{\textheight}{-2.4cm}   
                                  
In training across such a large range of friction values we found a divide arising between learning to perform the task reliably for a small range of friction values versus adequate performance across all of the friction values; since across all evaluation environments these two options yield the same total rewards. In theory, over longer training times, to maximize the rewards the agent should learn a policy that performs well in all environments.  

Additionally, anecdotally, we noticed the following behavior emerge: for larger table sizes where the vehicle could not reach sufficiently far with the fixed arm to manipulate the object, the agent would learn to push the object with the feet of the UAV and then fly back to view the object's progress before pushing with the feet again. In future work, varying the parameters of the training environments further could yield unique solutions to complicated problems that are often out of scope for and/or costly to develop with traditional controllers.

\section{Conclusion}

Our proposed approach will allow for more efficient and effective planning for a system in a complex environment with unknown dynamical properties. We expect this approach to be more robust to interacting with the environment (e.g. manipulation through foliage) and occlusions than previous approaches, since standard methods are frequently based on building a rigid geometric model of the environment and thus disallow any amount of contact with the environment. This work will enable a wide range of operational scenarios that are not currently feasible with existing technologies and will add to robustness in areas that are currently feasible. Ultimately, allowing robots to accomplish tasks that can be dangerous or inefficient for humans or in areas that a human can not reach.


\section*{Acknowledgement}
We gratefully acknowledge the support of
the National Science Foundation under grant \#1925189.

\bibliographystyle{IEEEtran}
\bibliography{references}

\end{document}